\documentclass[letterpaper, 10 pt, conference]{ieeeconf}  % Comment this line out if you need a4paper

\IEEEoverridecommandlockouts                              % This command is only needed if 
\usepackage{graphicx} % for pdf, bitmapped graphics files
\usepackage{amsmath} % assumes amsmath package installed
\usepackage{amssymb}  % assumes amsmath package installed
\usepackage{cite}
\usepackage{textcomp}
\usepackage{tabularx}
\usepackage{gensymb}
\usepackage{breqn}
\usepackage{xcolor}
\usepackage{placeins}
\usepackage{fancyhdr}

\usepackage[]{multirow}
\usepackage{subcaption}
\usepackage{tikz}
\usepackage{geometry}
\newcommand{\breakingcomma}{%
	\begingroup\lccode`~=`,
	\lowercase{\endgroup\expandafter\def\expandafter~\expandafter{~\penalty0 }}}

\newcommand*\titleheader[1]{\gdef\@titleheader{#1}}

\newcommand\copyrighttext{%
	\footnotesize © 2021 IEEE.  Personal use of this material is permitted.  Permission from IEEE must be obtained for all other uses, in any current or future media, including reprinting/republishing this material for advertising or promotional purposes, creating new collective works, for resale or redistribution to servers or lists, or reuse of any copyrighted component of this work in other works.}
\newcommand\copyrightnotice{%
	\begin{tikzpicture}[remember picture,overlay]
		\node[anchor=south,yshift=10pt] at (current page.south) {\fbox{\parbox{\dimexpr\textwidth-\fboxsep-\fboxrule\relax}{\copyrighttext}}};
	\end{tikzpicture}%
}

\title{\LARGE \bf
	Virtual Adversarial Humans finding Hazards in Robot Workplaces
}

\titleheader{2016 IEEE 24th International Requirements Engineering Conference}

\author{Tom P. Huck, Christoph Ledermann, and Torsten Kr\"oger% <-this % stops a space
	\thanks{This paper was submitted to the 2021 IEEE International Conference on Robotics and Automation (ICRA) and has been accepted for presentation. Please note that the final published paper may differ from this preprint version.}
	\thanks{This research was funded by the Ministry of Economics, Labor and Housing of the State of Baden-Württemberg in the research project 'RoboShield'}% <-this % stops a space
	\thanks{The authors are with the Intelligent Process Automation and Robotics Lab, Institute of Anthropomatics and Robotics (IAR-IPR), Karlsruhe Institute of Technology (KIT), 76131 Karlsruhe, Germany. Corresponding author: Tom Huck
		({\tt\small tom.huck@kit.edu})}%
}

\begin{document}

	\maketitle
%	\confheader{%
%		5$^{th}$  IEEE International Conference on Recent Advances and Innovations in Engineering - ICRAIE 2020 (IEEE Record\#51050)}
	\thispagestyle{empty}
	\pagestyle{empty}
	\copyrightnotice

	\begin{abstract}
	During the planning phase of industrial robot workplaces, hazard analyses are required so that potential hazards for human workers can be identified and appropriate safety measures can be implemented. Existing hazard analysis methods use human reasoning, checklists and/or abstract system models, which limit the level of detail.
	We propose a new approach that frames hazard analysis as a search problem in a dynamic simulation environment. Our goal is to identify workplace hazards by searching for simulation sequences that result in hazardous situations.
	We solve this search problem by placing virtual humans into workplace simulation models. These virtual humans act in an adversarial manner: They learn to provoke unsafe situations, and thereby uncover workplace hazards.
	Although this approach cannot replace a thorough hazard analysis, it can help uncover hazards that otherwise may have been overlooked, especially in early development stages. Thus, it helps to prevent costly re-designs at later development stages. For validation, we performed hazard analyses in six different example scenarios that reflect typical industrial robot workplaces.
	\end{abstract}
	
	%%%%%%%%%%%%%%%%%%%%%%%%%%%%%%%%%%%%%%%%%%%%%%%%%%%%%%%%%%%%%%%%%%%%%%%%%%%%%%%%
	\section{INTRODUCTION}
	\label{sec:Introduction}
	In recent years, there has been a trend in industrial automation towards human-robot collaboration \cite{Bauer2016}. Traditionally, safety issues with industrial robots were resolved by physically separating robot and human worker. In collaborative robotics this is often not possible. Instead, various combinations of different safety measures including detection sensors (e.g. laser scanners, light curtains), (partial) fencing, and software-based safety functions (e.g. velocity limitation, collision detection) are used, depending on the application and its required degree of collaboration. Flaws in the configuration of these safety measures can lead to hazards for human workers. To foresee potential hazards before commissioning, a hazard analysis is required \cite{ISO10218-2}. Currently, hazard analyses are based on experience, expert knowledge \cite{ISO14121}, and simple tools such as checklists (e.g. in \cite[Annex A]{ISO10218-2} and \cite[Annex B]{ISO12100}).\newline With the trend towards collaborative robotics, hazard analyses become more challenging due to greater system complexity. Thus, development of hazard analysis tools is an active field of research.
	\begin{figure}
		\includegraphics[width=1\columnwidth]{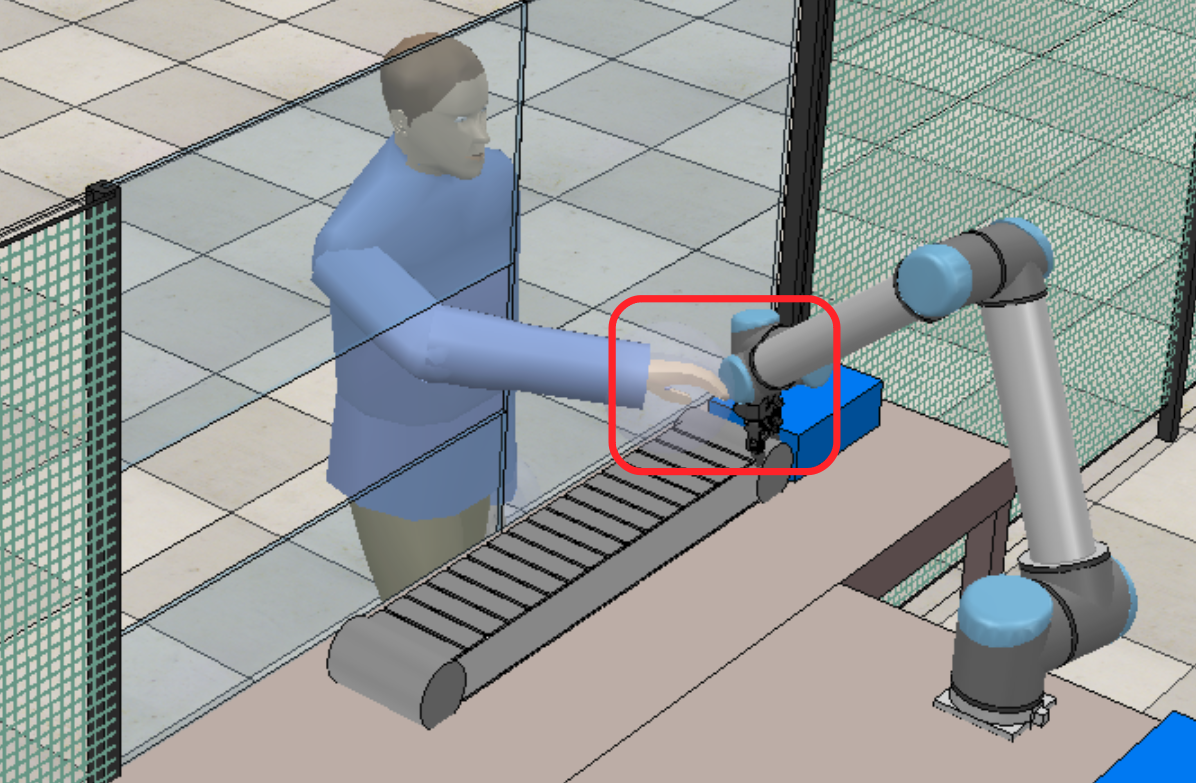}
		\caption{\small The virtual human autonomously explores the workplace to uncover hazards. By receiving rewards for high-risk behavior, it learns to actively provoke unsafe states. In this situation, for example, it reaches through a feed opening to provoke a collision.}
		\label{fig:ExampleSituation}
	\end{figure}
	Tools proposed in recent years typically use semi-formal or formal system models, e.g. control structure diagrams \cite{Leveson2016}, Product-Process-Resource models \cite{Awad2017}, UML models \cite{Guiochet2016}, or system descriptions in a formal language \cite{Askarpour2016,Vicentini2019}. These models have in common that they have a relatively \textit{high level of abstraction}, that is, system properties and dynamics are simplified to make complexity manageable. This has two major drawbacks:
	\begin{itemize}
		\item The level of detail is limited. Especially geometric aspects (e.g. cell layout, detection zones, safety fences) and dynamic aspects (e.g. collision forces, robot stopping time) require precise modeling, and thus, are difficult to consider in abstract system models.
		\item Common simulator models (CoppeliaSim, Process Simulate, etc.), which are often already available in the development process, are not used. Instead, new purpose-specific models are obtained which increases the cost and time needed for hazard analysis.
	\end{itemize}
	Considering these drawbacks, there is a need for a hazard analysis tool that operates directly in a dynamic simulation model (i.e. on a \textit{low abstraction level}) without the need to first derive another model that is abstracted or simplified.\newline
	A major challenge of such a simulation-based hazard analysis is to create conditions where hazards are uncovered: In complex simulation models the number of possible simulation sequences can be vast and an exhaustive search can be infeasible. If certain hazards only manifest themselves in very specific simulation sequences, these hazards can remain undiscovered. We address this problem by introducing a virtual human into the simulation that acts in a manner adversarial to the safety measures: It learns to deliberately provoke unsafe situations, which increases the chance of uncovering hazards. We initially proposed this idea in \cite{Huck2020}. In this paper, we further develop the idea and present a concept for a simulation-based hazard analysis tool. In particular, we extend our work from \cite{Huck2020} with a body-region-specific collision-force estimation to evaluate hazards with respect to the potential collision severity. Furthermore, we use an articulated arm model to determine reachability of hazardous areas. This enables the finding of hazards that result from reaching behind or around safety barriers (compare Fig. \ref{fig:ExampleSituation}). Our conceptual tool was evaluated by testing it on six different industrial robot workplaces with different safety configurations and hazards. 
	\section{RELATED WORK AND STANDARDS}
	\label{sec:RelatedWork}
	\subsection{Tools and Methods for Hazard Analysis}
	\label{sec:MethodsAndTools}
	In industrial practice, hazard analysis is usually conducted according to the \mbox{ISO 12100} standard \cite{ISO12100}. However, the procedures recommended in \mbox{ISO 12100} are quite generic and do not address the specific challenges arising for robot systems. Thus, several novel tools and methods have been proposed specifically for the analysis of robot systems.\newline
	\textbf{Semi-formal:} Semi-formal analysis methods like “Hazard and Operability Analysis” (HAZOP) \cite{IEC61882} or “Systems-Theoretic Process Analysis” (STPA) \cite{Leveson2016} have been adapted to analyze robot systems. These methods are based on semi-formal system descriptions, e.g. flow diagrams (HAZOP) or control structure diagrams (STPA). Guiochet proposed a HAZOP-Extension that uses UML diagrams for the analysis of robot systems \cite{Guiochet2016}. Marvel et al. have proposed a method that is based on an ontology of basic human-robot collaborative tasks \cite{Marvel2014}.%All of these methods have in common that thy define certain procedures to guide the user through the analysis. However, the actual hazard identification is still based on human reasoning and can place a high analytical burden on the user, especially when analyzing complex systems.
	\newline
	\textbf{Formal and Rule-based: } Awad et al. have developed an expert system for automated risk assessment of robot workplaces \cite{Awad2017} using a model of processes, products and resources (PPR-model). A rule base maps the PPR-model to hazards. This approach was further developed by Wigand et al \cite{Wigand2020}. The method "SAFER-HRC", developed by Askarpour et al. \cite{Askarpour2016,Askarpour2017}, and Vicentini et al. \cite{Vicentini2019}, uses formal verification for hazard analysis. Based on a system description in the formal language TRIO, a bounded satisfiability checker is used to verify safety properties of collaborative robot systems.\newline%\textbf{[TODO: Verweis auf Caterino et al. (Petri Nets])}\newline
	\textbf{Simulation-based:} In this paper, we propose a simulation-based approach. The use of simulation for analysis of safety-critical systems is well-researched \cite{Corso2020}. However, in the domain of robot safety, simulation-based approaches are usually employed on a component- rather than a system-level, for example to verify control code or control strategies \cite{Araiza2015,Araiza2016,Uriagereka2019, Bobka2016}. Askarpour et al. recently presented a co-simulation approach which couples SAFER-HRC with the simulator MORSE \cite{Askarpour2020}. However, MORSE is used primarily for visualization, and the underlying hazard identification is still conducted with a formal model. %To the best of our knowledge, there are currently no other simulation-based approaches for hazard analysis of robot workplaces.
	\subsection{Safety Standards in Industrial Robotics}
	\label{sec:Standards}
	For a better understanding of the following contents, we will shortly introduce relevant safety standards besides the already mentioned \mbox{ISO 12100}. Robot safety is governed by the standard \mbox{ISO 10218} \cite{ISO10218-1,ISO10218-2} and the technical specification \mbox{ISO/TS 15066}\cite{ISOTS15066}. Here, especially \mbox{ISO/TS 15066} is important. It defines body-region specific force and pressure limits for human-robot collisions. Safety measures must either ensure that collisions are avoided, or that collision force and pressure remain below these limits. \mbox{ISO/TS 15066} Annex A provides a calculation model with body-region specific parameters to estimate collision forces. Other relevant standards are \mbox{ISO 13855}\cite{ISO13855} and \mbox{ISO 13857}\cite{ISO13857}. \mbox{ISO 13855} regulates the use of contactless safeguards (e.g. laser scanners, light curtains) which are used to detect human workers. \mbox{ISO 13857} regulates the placement and dimensions of safety fences and ensures that hazardous areas cannot be accessed by reaching over or around the fences.
	
	\section{{OUR APPROACH}}\label{sec:Approach}
	In contrast to the (semi-)formal and rule-based methods from Section \ref{sec:MethodsAndTools}, our goal is a purely simulation-based approach to hazard analysis. We focus on finding hazards that result from systematical faults in the design, configuration, or programming of the robot system. We do not consider hazards that result from random faults (e.g. component failures), because these types of hazards are typically addressed on a component- rather than a system level, and there are already established safety engineering methods to regard them (e.g. Failure Mode, Effects and Diagnostics Analysis, FMEDA \cite{Goble1999}, or Fault Tree Analysis, FTA \cite{IEC61025}).	Under these premises, we assume that for a given human action, the reaction of the robot system is deterministic. If the system has inherent hazards, then there will be sequences of human actions that can provoke hazardous situations. Thus, the problem of hazard analysis can be framed as a search problem, where the goal is to find sequences of human actions that result in a hazardous situation. Formally, this approach can be described as follows:
	\begin{equation}
	\langle S,U,A,\phi, s_0 \rangle
	\end{equation}
	where $S$ is a set of simulator states $s$ describing the configuration of human, robot, and workplace environment and $A$ is a set of actions which can be performed by the virtual human in simulation. $U$ is a subset of $S$ representing unsafe states (i.e. hazardous situations in simulation). We define an unsafe state as a state where there is contact between human and robot while the robot is moving with a velocity that could potentially lead to a collision force exceeding the \mbox{ISO/TS 15066} limits:	
	\begin{equation}
	U=\left\{s\ |\ d_{HR} = 0,\ F_{\mathrm{coll}}>F_{\mathrm{max}}\right\}
	\end{equation}
	Here, $d_{HR}$ is the human-robot distance, $F_{\mathrm{coll}}$ the potential collision force estimated according to \mbox{ISO/TS 15066} equations (A.1)-(A.4)\footnote{In this paper, we estimate $F_{coll}$ based on the \textit{absolute} velocity of the robot at contact point; the \textit{direction} is not considered, because our simplified human model does not allow for reliable calculation of the collision vector. Thus, our definition (2) is conservative: It also applies to contacts where the robot velocity points away from the human (compare Video, Scenario 6). This limitation will be addressed in the future with more detailed modeling.}, and $F_{\mathrm{max}}$ denotes the body-region-specific limit (see \mbox{ISO/TS 15066}, Annex A).
	%%%%%%%%%%%%%%%%%%%%%%%%%%%%%%%%%%%%
	The function $\phi$ is a transition function that returns the next state given the current state and a human action: $s_{k+1}=\phi(s_k,a)$. This function is implemented by the simulator: The next state is obtained by simulating the behavior of human and robot system for a given human action.\newline
	The search problem now consists of finding sequences of human actions $a_1, a_2, ... a_n$
	that, when performed in interaction with robot and workplace environment, result in an unsafe state $s \in U$.
	Depending on the number $|A|$ of possible actions and the length $n$ of the sequences, the search space, which consists of $|A|^n$ possible simulations sequences, can be vast. To address this challenge, we combine the following methods:
	\begin{itemize}
		\item \textbf{Monte Carlo Tree Search (MCTS):} We use MCTS to search for human action sequences that result in critical collisions. By rewarding the MCTS algorithm for the occurrence of dangerous situations, the virtual human learns to deliberately provoke unsafe states. This increases the chance of uncovering hazards and thus mitigates the problem of large search spaces \cite{Lee2018,Huck2020}.
		\item \textbf{Reachability Analysis:} To reduce the human action space, we combine MCTS with a reachability analysis. In particular, we use reachability analysis of the human arm to determine if the robot is reachable by the human. We thus avoid modeling the complex arm movements of humans in the form of explicit actions. This prevents an explosion of the search space.
	\end{itemize}
	
	\begin{figure}[t]
		\centering
		\includegraphics[width=1\columnwidth]{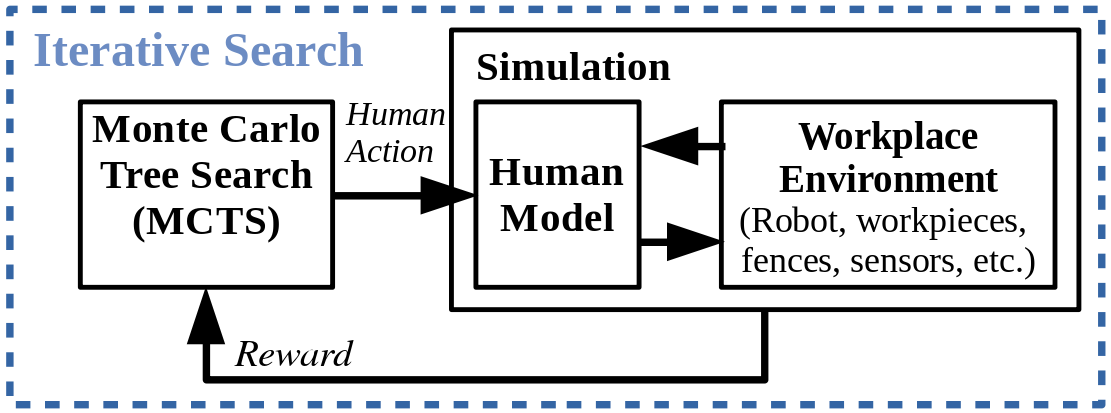}
		\caption{ \small The MCTS algorithm selects a sequence of human actions which the human model then executes in the simulation. Based on the level of danger in the current simulation state, the algorithm receives a reward that is designed to encourage dangerous behavior and thus increases the chance of finding hazards.}
		\label{fig:SystemOverview}
	\end{figure}
	
	\section{IMPLEMENTATION}
	\label{sec:Implementation}
	\begin{figure}
		\centering
		\includegraphics[width=0.7\columnwidth]{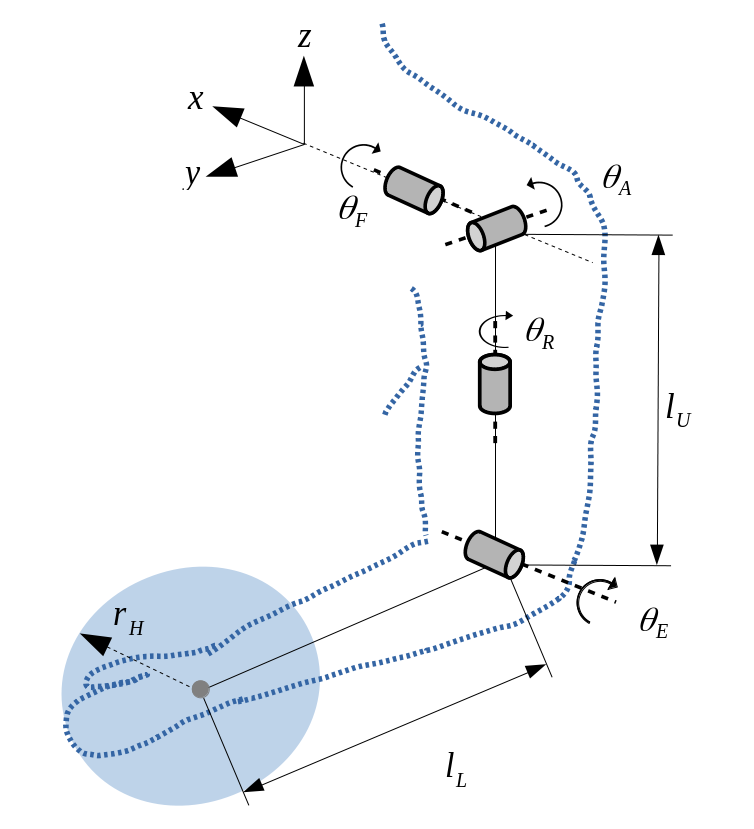}
		\caption{\small 4-DoF arm model with $\theta_F$ for flexion and $\theta_A$ for abduction of the elbow, $\theta_R$ for upper arm rotation, and $\theta_E$ for elbow flexion. Lower arm rotation and wrist flexion are omitted. Instead, a bounding sphere is attached to the wrist, over-approximating the possible hand poses. Upper arm length $l_U$, lower arm length $l_L$, and radius of the bounding sphere $r_H$ are parameterized according to \mbox{ISO 13857}.}
		\label{fig:ArmModel}
	\end{figure}
	Our implementation is based on the iterative search procedure shown in Fig. \ref{fig:SystemOverview}: The MCTS algorithm selects a human action which is then executed by the human model in interaction with the simulated workplace environment.
	After each search step the level of danger in the current simulation state is evaluated. Based on the danger level the MCTS algorithm receives a reward which it then uses to adapt the search behavior. When an unsafe state is found, or a maximum number of search steps is reached, the simulation is set back to the initial state and the next search iteration begins. The remainder of this section describes the human model, the MCTS algorithm, and the reward in more detail.\newline
	\textbf{Human Model. } We use "Bill", a simple human model included in the simulator CoppeliaSim (former V-REP \cite{Rohmer2013}). The model is adapted so that it can perform six walking- and six upper-body-motions, amounting to an action space of 36 combined motions:
	\begin{dmath}\breakingcomma
		A_{\mathrm{Walking}}=\left\{(\mathrm{stop}), (\mathrm{forward}),(\mathrm{left\ 45\degree}),(\mathrm{left\ 90\degree}),(\mathrm{right\ 45\degree}),(\mathrm{right\ 90\degree}) \right\}
		\label{eq:ActionSpaceStart}
	\end{dmath}
	\begin{dmath}\breakingcomma
		A_{\mathrm{UpperBody}}=\left\{ (\mathrm{upright}),(\mathrm{bend\ forward)},(\mathrm{bend\ left)},\mathrm{(bend\ right)},\mathrm{(bend\ forward\ and\ right)},\mathrm{(bend\ forward\ and\ left)} \right\}
	\end{dmath}
	\begin{gather}
	\label{eq:HumanActionEnd}
	A=A_{\mathrm{Walking}} \times A_{\mathrm{UpperBody}}\\
	|{A}|=6\cdot 6= 36
	\label{eq:ActionSpaceEnd}
	\end{gather}
	The walking speed is 1.6m/s, which is assumed as human approach speed in \mbox{ISO 13855} (compare Section \ref{sec:Standards}).\newline
	To prevent an explosion of the action space, arm motions are not modeled in the form of explicit actions. Instead, the virtual human performs a reachability check in each timestep to find potentially dangerous arm configurations.
	For the reachability check, we use an articulated model that allows the virtual human to also find collisions that require reaching around or behind protective barriers (see Fig. \ref{fig:ExampleSituation}). The arm model, seen in Fig. \ref{fig:ArmModel}, is adapted from \cite{Klopcar2005} to replicate the arm model on which \mbox{ISO 13857} is based \mbox{\cite[Table 3]{ISO13857}}. It features joints for shoulder abduction and flexion, upper arm rotation, and elbow flexion. Wrist flexion and lower arm rotation are omitted. To make up for the omission, a sphere is placed around the wrist for an over-approximate collision detection. In each timestep the virtual human checks if the robot is close enough to be potentially reachable. If this is the case, the reachability check is performed by sampling through the joint angle limitations given by \cite{Klopcar2005} and checking for collisions with robot and obstacles.
	
	\textbf{Algorithm. }For the MCTS algorithm we use a Monte Carlo Tree Search with Double Progressive Widening (MCTS-DPW) from Lee et al. \cite{Lee2018}. For reasons of brevity, we only describe the algorithm in a simplified manner. For a detailed explanation, we refer to \cite{Lee2018}.
	\begin{figure}[t]
		\centering
		\includegraphics[width=1\columnwidth]{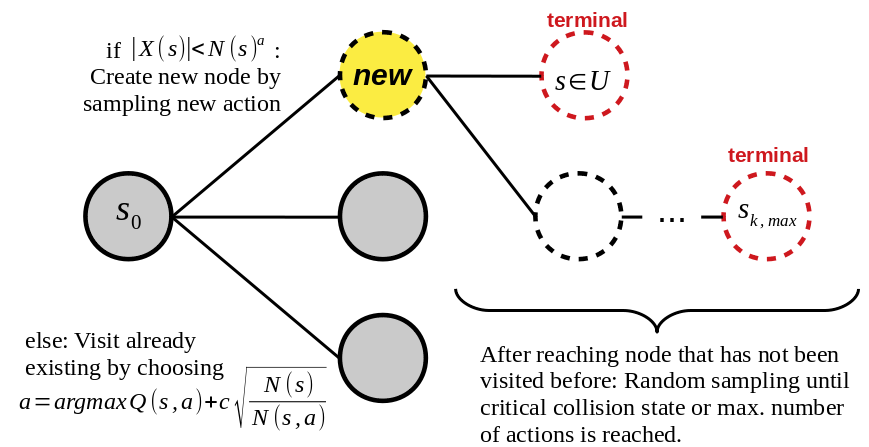}
		\caption{ \small Search tree: Each search iteration starts in the initial simulation state $s_0$. At each node the algorithm decides whether to add a new node to the search tree by randomly sampling a new action, or re-visit an existing node by selecting an action based on the state-action values estimate. When the algorithm reaches a new node which has not been visited before, an action sequence is sampled randomly until it terminates by reaching an unsafe state $s \in U$ or by reaching the maximum number of search steps $k_{max}$}
		\label{fig:SearchTree}
	\end{figure}
	
	The algorithm iteratively samples action sequences and executes them in the simulator. While doing so, it builds a search tree (see Fig. \ref{fig:SearchTree}) where the nodes correspond to states and the edges to human actions\footnote{Technically, the implementation of Lee et al. selects actions indirectly via adjusting the random seed of the simulator. In our case, however, each seed corresponds to one specific action. Thus, for a clearer notation, we denote the action directly.}. 
	While stepping through the search tree, a state-action value function $Q(s,a)$ is estimated to guide the search. 
	A search iteration is started by initializing the simulator with the initial state $s_0$. The algorithm then steps through the search tree by selecting human actions and executing them in the simulation. In each search step, the algorithm decides whether to add a new node by sampling a new action from a uniform distribution over $A$, or whether to re-visit an existing node. A new node is added if the following criterion holds:
	\begin{equation}
	|X(s)|<k\cdot N(s)^\alpha
	\label{eq:progressiveWideningCriterion}
	\end{equation}
	Here, $|X(s)|$ is the number of different actions that have already been executed in this state (i.e. the number of edges starting at the current node), $N(s)$ is the number of times the node has already been visited, and $k,\alpha$ are parameters to balance between explorative and exploitative search behavior.s
	If (\ref{eq:progressiveWideningCriterion}) does not hold, an existing node is revisited by choosing an already explored action based on the estimated $Q(s,a)$:
	\begin{equation}
	a=\mathrm{argmax}\left\{Q(s,a)+c\cdot \sqrt{\frac{N(s)}{N(s,a)}}\right\}
	\label{eq:seedSelection}
	\end{equation}
	Here, $N(s,a)$ is the number of times the action $a$ was executed in state $s$, and $c$ is another parameter to balance exploration and exploitation.\newline
	Once the algorithm has reached a new node that has not been visited before, actions are sampled randomly from a uniform distribution until the simulation terminates. The simulation terminates when an unsafe state $s \in U$ occurs, or when the maximum number of search steps $k_{max}$ is reached. After the simulation terminates, it is set back to initial state $s_0$ and the next iteration of the search is started.\newline
	The algorithm receives a reward $R$ after each action. When reaching a terminal state, the state-action value function is updated via backpropagation:
	\begin{equation}
	Q(s,a)\leftarrow Q(s,a)+\frac{q-Q(s,a)}{N(s,a)}
	\end{equation}
	where $q$ is the undiscounted sum of rewards that have been incurred in previous nodes of the backpropagation sequence.\newline
	\textbf{Reward. }The reward $R$ is defined as follows:
	\begin{equation}
	R = \begin{cases}
	R_E & \text{if}\ s \in U\\
	-\frac{1}{c_D} & \text{if}\ s \notin U\ \mathrm{and}\ k=k_{\mathrm{max}}\\
	0 & \text{otherwise}
	\end{cases}
	\label{eq:Reward}
	\end{equation}
	Where the first component $R_E$ is a non-negative constant to reward the occurence of an unsafe state and the second component is a negative penalty that is given if the maximum number of actions $k_{max}$ is reached without finding an unsafe state. The penalty is based on $c_D$, which characterizes the level of danger in a given simulation state. Since we want to penalize \textit{safe} states (i.e. states with a small $c_D$) higher, we give the \textit{inverse} of $c_D$ as penalty.\newline
	The danger level $c_D$ is calculated by a heuristic function that differentiates between three cases:
	\begin{equation}
	c_D = \begin{cases} \frac{F_{\mathrm{coll}}}{F_{\mathrm{max}}} & \text{in case (a)} \\
	\min \left\{\frac{F_{\mathrm{coll,virt}}}{F_{\mathrm{max}}}*e^{-d_{\mathrm{HR}}},1\right\} & \text{in case (b)}\\
	0.01 & \text{in case (c)}
	\end{cases}
	\label{eq:dangerIndex}
	\end{equation}
	\begin{itemize}
		\item[(a)] Human-robot contact ($d_{HR}=0$).
		\item[(b)] Human and robot are close, but no contact ($d_{HR}\leq$1.5m).
		\item[(c)] Human-robot distance is larger than 1.5m or human and robot are fully separated by a barrier.
	\end{itemize}
	\indent In \textit{case (a)}, the danger level is defined as the ratio between the potential collision force and the maximum acceptable collision force for the affected body part.\newline
	\indent In \textit{case (b)}, where there is no direct human-robot contact, the danger level is calculated on the basis of a virtual collision force $F_{coll,virt}$, which is the potential collision force that would occur if the closest human-robot object pair would collide. To take distance into account, $c_D$ is scaled down with the human-robot distance $d_{\mathrm{HR}}$. Note that $c_D$ is limited to 1, so that a situation from case (b) always returns a lower value than an actual contact situation exceeding the limit.\newline
	\indent In \textit{case (c)}, we assume that the level of danger is negligible, and therefore define $c_D=0.01$ ($c_D > 0$ to avoid division by zero in (\ref{eq:Reward})).\newline
	\textbf{Postprocessing: }During the search, unsafe states are documented in a log file that contains the potential collision force, the affected body region, and the sequence of actions that has led to the unsafe state. Based on the log information, the user can re-play the simulation sequence and decide if the unsafe state constitutes a realistic hazard that requires further safety measures. It is important to note that this final decision still needs to be made by a human: Our method is primarily intended to raise awareness of \textit{possible} hazards, but due to modeling simplifications or an overly conservative definition of unsafe states, not all hazards that are possible in simulation are necessarily realistic in the context of a real-world application.
	
	\section{APPLICATION EXAMPLES}
	\label{sec:ApplicationExample}
\begin{figure}
		\includegraphics[width=\columnwidth]{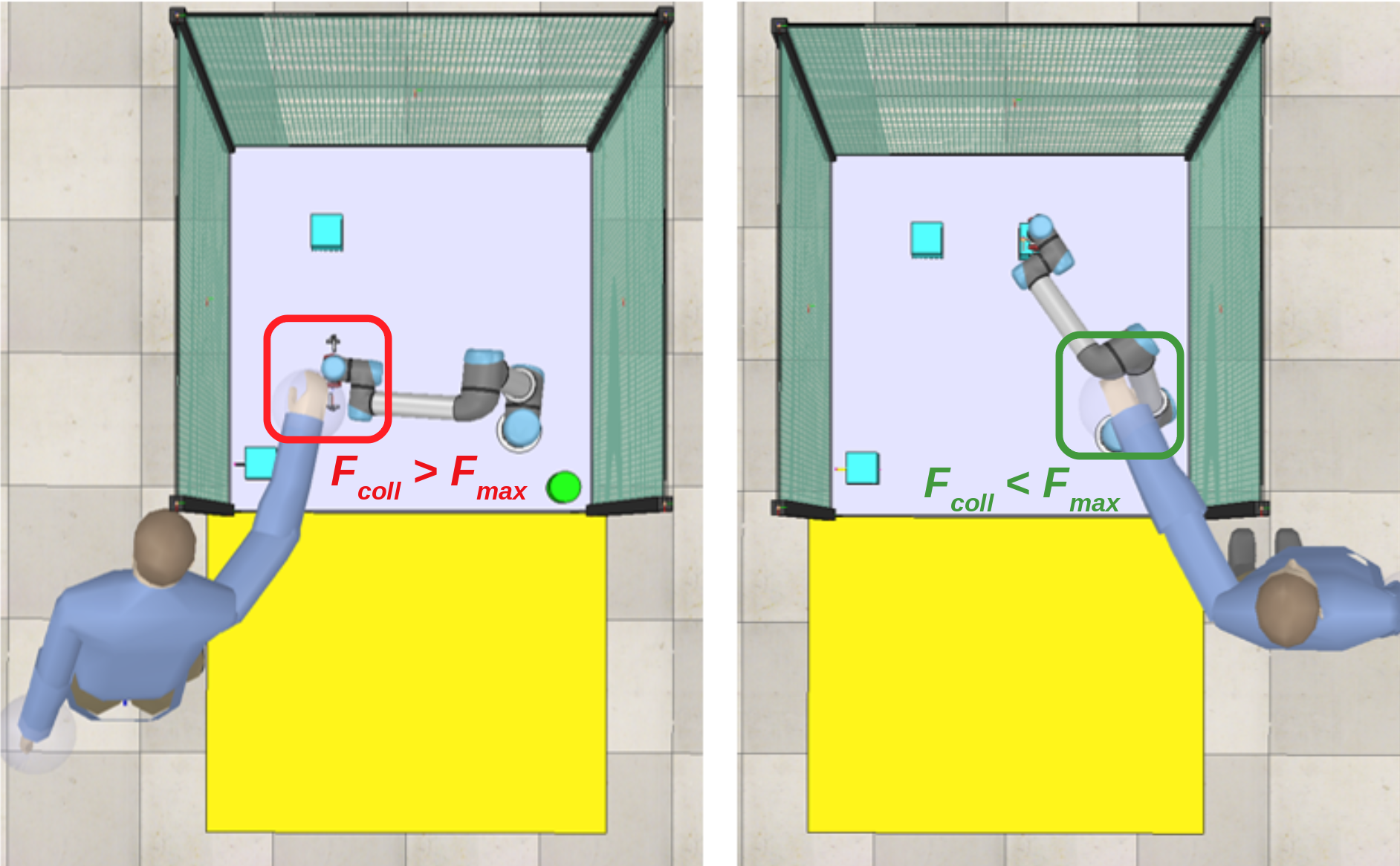}
		\caption{\small Example from Scenario 2: In both situations, the human provokes a collision by avoiding the safety mat (yellow) and reaching around the fence. However, only on the left side, the potential collision force exceeds the \mbox{ISO/TS 15066} limit. The situation on the right side is a local optimum w.r.t the reward function, i.e. collisions are possible, but do not represent an unsafe state according to our definition (2).}
		\label{fig:localMinimum}
	\end{figure}

	We demonstrate our approach in six different test scenarios that reflect typical industrial robot workplaces\footnote{To avoid the scenarios becoming too convenient due to bias (i.e. hazards being too easy to find), they were designed and implemented by persons who were not involved in the development of the proposed method.}. Each scenario has a deliberate safety flaw that results in a specific hazard. We deployed our virtual human as described in Section \ref{sec:Implementation} to find these hazards. As a baseline we also deployed a variant that uses a random search instead of the MCTS algorithm. In this variant, the human action sequences are sampled randomly from a uniform distribution over $A$ (the other functions, including the reachability check, remain unchanged). For each combination of scenario and search algorithm, we conducted ten test runs with different random seeds. Each search iteration consists of a sequence of 8 actions, each with a duration of 200ms. Length of the test runs was limited to 320 iterations.\newline
	The test scenarios are explained in Table I and in the video accompanying this paper. For each test scenario and each algorithm, we calculated the average runtime $T$ (i.e the average number of search iterations needed to find the hazard), and the miss rate $N_{miss}$ (i.e. in how many of the ten test runs the hazard was not found). These results are also shown in Table I. Both MCTS and random search successfully identified the hazards in the majority of test runs. However, in case of the random search, there were seven instances of hazards being missed, whereas MCTS only missed a hazard in one instance. Furthermore, compared to the random search, the average runtime of MCTS was significantly smaller in all scenarios. These results indicate that adapting the human behavior to provoke more dangerous situations does indeed increase the chance of finding hazards.\newline
	We also conducted a closer investigation of the one instance where MCTS missed a hazard. We found that this was due to premature convergence to a local optimum, that is, a collision where the reward is relatively high, but the situation does not constitute an unsafe state in the sense of our definition (2). Convergence to such a local optimum can prevent the algorithm from discovering a more critical situation (compare Fig. \ref{fig:localMinimum}).
	
	\begin{table*}[!h]
		\begin{tabularx}{\textwidth}{|p{0.7\textwidth}|p{0.25\textwidth}|}
			\hline
			\textbf{Test scenario and hazard}	& \textbf{Results$^*$}\\
			\hline
			\textbf{1 Safety fence} (Fig. \ref{fig:ExampleSituation}): 			
			The robot operates behind a safety fence and plexiglass barrier. The plexiglass barrier has a feed opening which provides access to a conveyor belt. 
			The safety distance between the feed opening and the robot path is insufficient. An unsafe state is possible when the worker reaches behind the feed opening and collides with the robot.
			& \textbf{MCTS:}\newline $T = 16.3,\ \sigma=12.8,\ N_{miss}=0$\newline
			\textbf{Random Search:}\newline $T = 37.4,\ \sigma=52.8,\ N_{miss}=0$\\
			\hline
			\textbf{2 Sensor mat } (Fig. \ref{fig:localMinimum}): 
			The robot works in a cell which is encased by fences on three sides, but has an open front. A sensor mat in front of the cell stops the robot when the worker steps on it. However, the mat is not wide enough. It only covers the front of the cell, not the right and left side. Thus, an unsafe state can occur when the worker walks around the mat and reaches behind the fence into the cell.
			& \textbf{MCTS:}\newline $T = 131.6,\ \sigma=76.6,\ N_{miss}=1$\newline
			\textbf{Random Search:}\newline $T = 203.4,\ \sigma=133.4,\ N_{miss}=4$ \\
			\hline
			\textbf{3 Head collision } (Fig. \ref{fig:Hazards}-a) 
			The robot performs a pick-and-place task on a table. A laser scanner detects the approach of workers and reduces the robot velocity. With the reduction, collisions with arms and upper body are uncritical. However, due to the height of the table the robot can also collide with the head which is still critical despite the velocity limitation.
			&  \textbf{MCTS:}\newline $T = 83.1,\ \sigma=45.8,\ N_{miss}=0$\newline
			\textbf{Random Search:}\newline $T = 115.8,\ \sigma=113.0,\ N_{miss}=1$ \\
			\hline
			\textbf{4 Light curtain } (Fig. \ref{fig:Hazards}-b) The worker can enter the robot cell through a light curtain. Upon entrance the robot stops with a certain stopping time. If the worker enters while the robot passes a certain point on its path, a part of robot protrudes far enough out so that the worker can reach it before the robot has slowed down sufficiently, resulting in an unsafe state.
			& \textbf{MCTS:}\newline $T = 14.9,\ \sigma=13.4,\ N_{miss}=0$\newline
			\textbf{Random Search:}\newline $T = 28.5,\ \sigma=23.4,\ N_{miss}=0$  \\
			\hline
			\textbf{5 Laserscanners } (Fig. \ref{fig:Hazards}-c) 
			Two robots work in parallel. Each robot is safeguarded by a separate laser scanner zone (red, yellow). When a worker enters one of the zones, the respective robot stops while the other keeps running. However, an unsafe state is possible when the worker enters one of the laser scanner zones, but then reaches towards the robot on the opposite side, whose laser scanner field is not activated.
			& \textbf{MCTS:}\newline $T = 69.4,\ \sigma=66.8,\ N_{miss}=0$\newline
			\textbf{Random Search:}\newline $T = 174.3,\ \sigma=137.7, \ N_{miss}=2$  \\
			\hline
			\textbf{6 Heavy workpiece } (Fig. \ref{fig:Hazards}-d) 
			The robot reaches outside its workcell to grasp a heavy workpiece which it then takes inside. Inside the cell, the robot works with full velocity. Outside, the velocity is limited to keep collision forces below acceptable thresholds.
			However, the limitation does not consider the additional effect of the payload. As long as no workpiece is grasped, collisions are still uncritical. But if a contact occurs after the workpiece is grasped, the potential collision force exceededs the limits.
			& \textbf{MCTS:}\newline $T = 24.6,\ \sigma=25.6,\ N_{miss}=0$\newline
			\textbf{Random Search:}\newline $T = 71.1,\ \sigma=57.9,\ N_{miss}=0$   \\
			\hline
		\end{tabularx}
		\label{tab:TestScenarios}
		\caption{\small Test Scenarios and Results. $^*$Notation: $T$ is the runtime, i.e. the average number of search iterations needed to find the hazard. $\sigma$ is the standard devation of the runtime. $N_{miss}$ denotes the number of unsuccessful test runs where the respective hazard was missed.}
	\end{table*}
	
	\begin{figure*}
		\centering
		\begin{subfigure}[t]{0.222\textwidth}
			\includegraphics[width=1\textwidth]{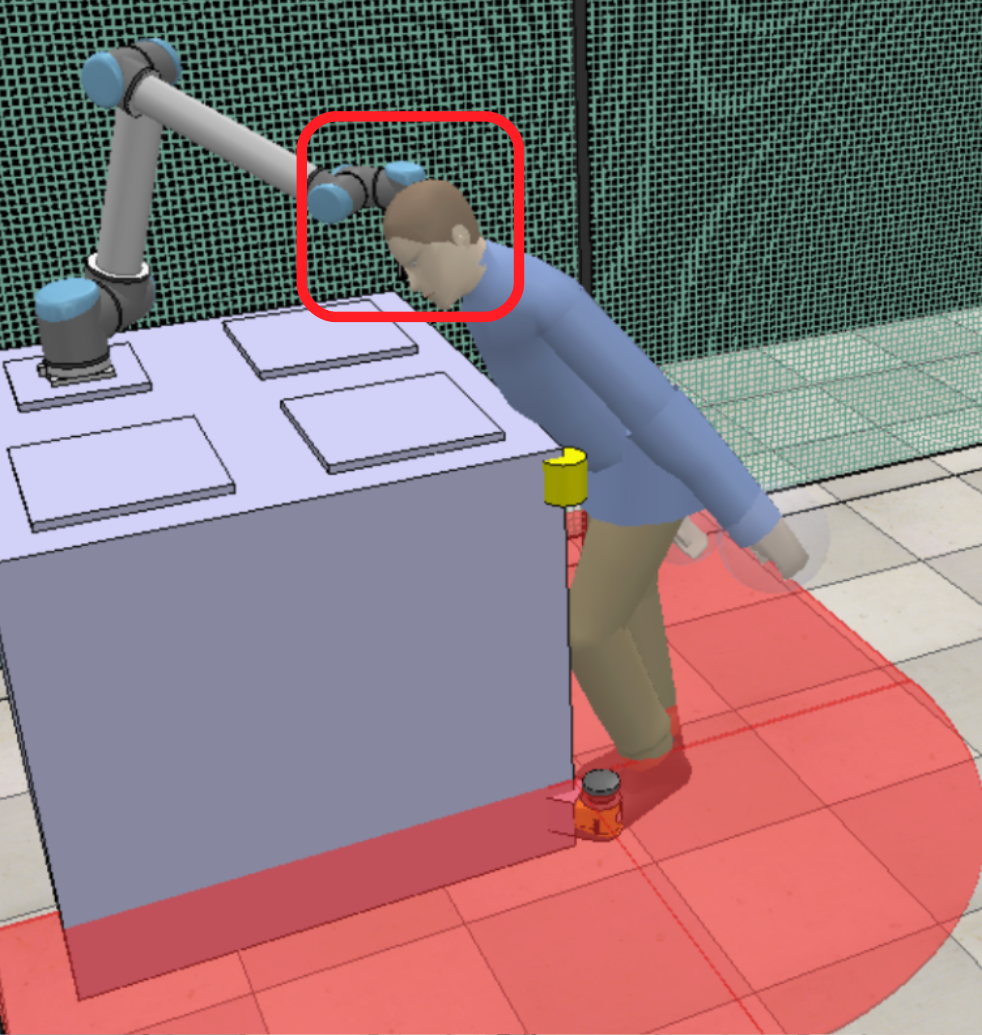}
			\caption{Scenario 3}
		\end{subfigure}
		\hfill
		\begin{subfigure}[t]{0.2\textwidth}
			\includegraphics[width=1\textwidth]{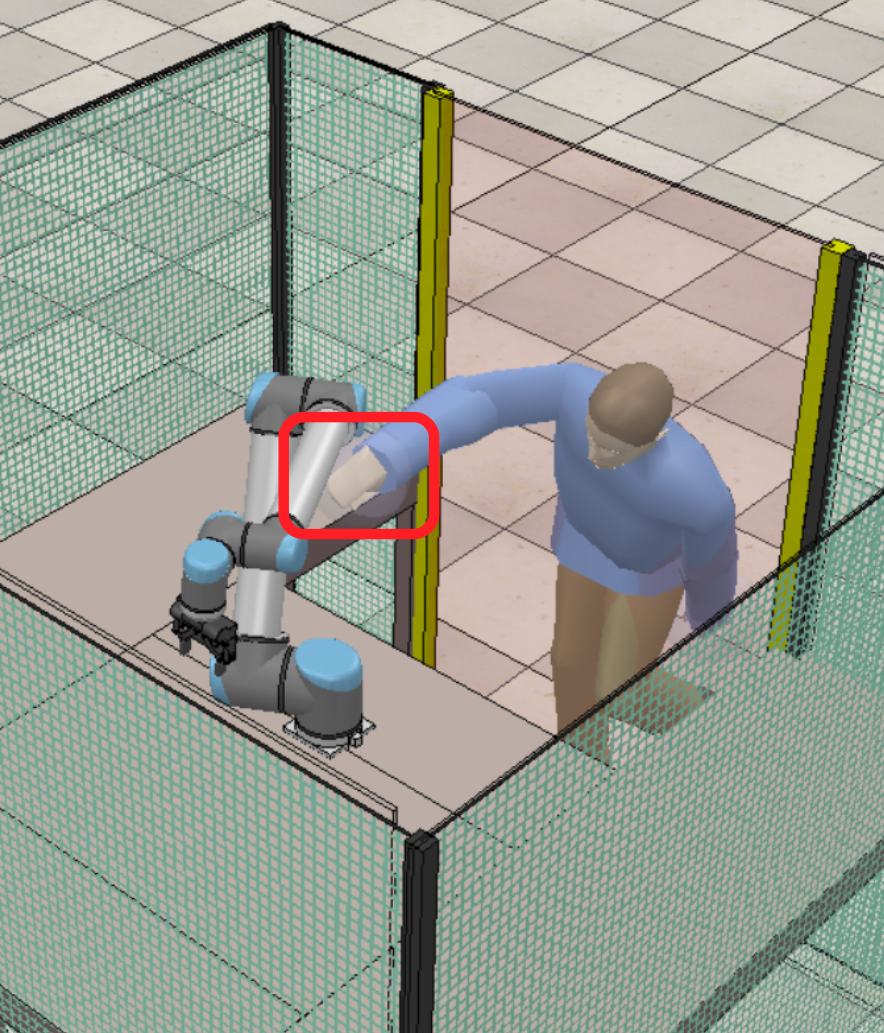}
			\caption{Scenario 4}
		\end{subfigure}
		\hfill
		\begin{subfigure}[t]{0.32\textwidth}
			\includegraphics[width=1\textwidth]{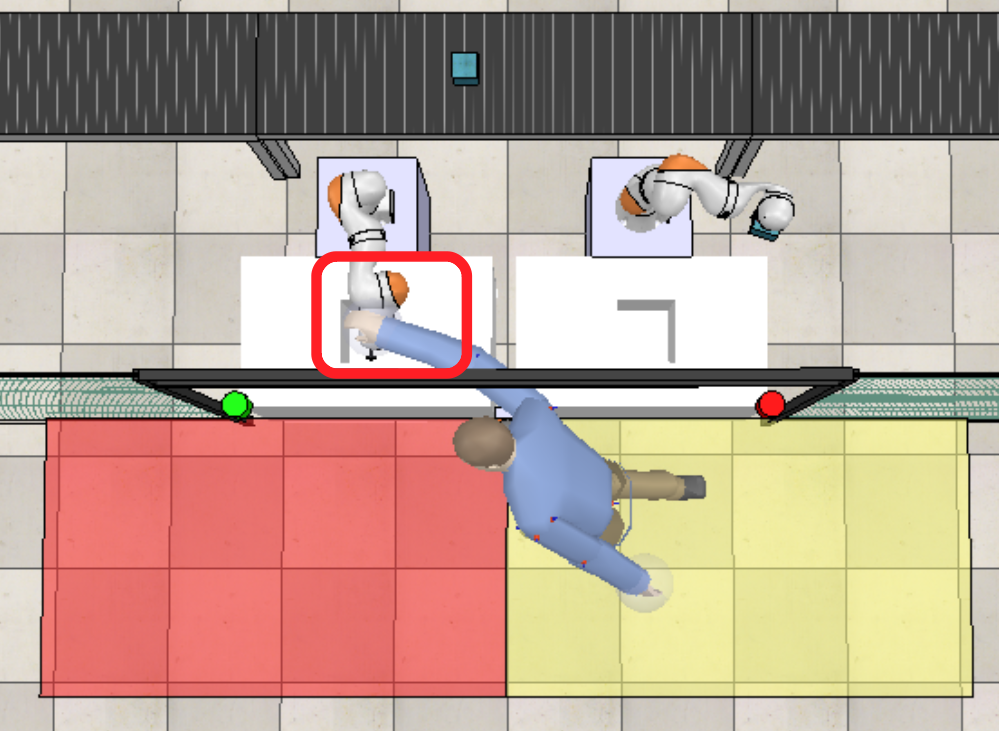}
			\caption{Scenario 5}
		\end{subfigure}
		\hfill
		\begin{subfigure}[t]{0.226\textwidth}
			\includegraphics[width=1\textwidth]{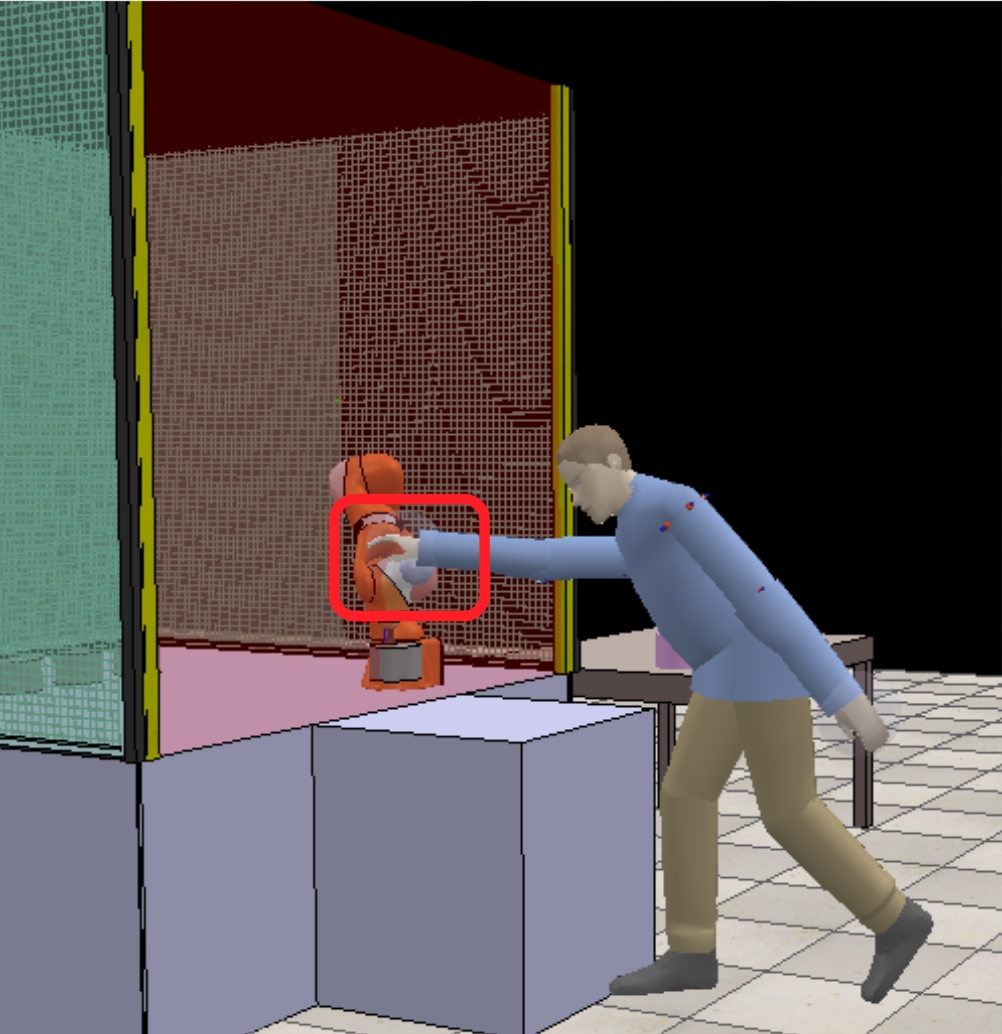}
			\caption{Scenario 6}
		\end{subfigure}
		\caption{Examples for hazardous situations found in Scenarios 3-6 (for scenarios 1 and 2, see Fig. \ref{fig:ExampleSituation} and Fig. \ref{fig:localMinimum}).}
		\label{fig:Hazards}
	\end{figure*}
	
	\section{DISCUSSION}
	\label{sec:Discussion}
	The success of our prototypical implementation indicates that our approach is feasible and can find hazards in realistic simulation models with a high success rate. The comparison to the random search also confirms that our approach of generating adversarial human behavior improves search performance.\newline
	Compared to other hazard analysis methods, which typically use simplified or abstracted system models, our approach has the benefit that it essentially treats the system model as a black-box (compare Fig. \ref{fig:SystemOverview}), which means that there is no inherent limit regarding the modeling principles or the model's level of detail. Thus, our method can be deployed on detailed and dynamic simulation models without extensive simplification or abstraction. Furthermore, the approach only requires little prior knowledge about the internal workings of the system that is analyzed, which means that it is easy to deploy for untrained users. In future applications, these properties will also be useful when analyzing systems that are highly complex or rely on black-box components that are difficult to assess with traditional methods (e.g. neural networks).\newline However, the black-box character also has disadvantages: In cases where prior knowledge is available, it cannot be incorporated into the search to increase efficiency. Also, the approach is based solely on falsification, that is, it can find situations where safety criteria are violated, but it cannot give a guarantee that a system is safe (which, however, is the case for all available hazard analysis methods, compare Sections \ref{sec:MethodsAndTools} and \ref{sec:FutureWork}).\newline
	Regarding limitations of the current implementation, it should be noted that the human action space $A$ is quite simplistic. Furthermore, the search algorithm is currently not suitable to deal with local optima or multiple hazards in one scenario (compare Section \ref{sec:ApplicationExample} and Fig. \ref{fig:localMinimum}).
	
	\section{CONCLUSION AND FUTURE WORK}
	\label{sec:FutureWork}
	Given the promising results in the test scenarios, we aim to develop our approach further into a hazard analysis tool that can be used in industrial practice. To achieve this, we will explore different options for a more detailed human model. Along with this, we will also improve the collision force estimation to improve accuracy and avoid overly conservative estimations (compare Section \ref{sec:Approach} and Footnote 1).
	Apart from these rather straightforward improvements, we must also address a number of more fundamental challenges:
	\begin{itemize}
		\item \textit{{Action-/Search Space:}} In our example, the action space only consists of basic movements. However, depending on the application, other, perhaps more complex actions can be relevant. This leads to the question what types of actions the human action space should contain and how they should be represented. Also, a more fine-granular action space is likely to present new challenges regarding computational complexity.\newline
		\item \textit{Prior Knowledge:} How can we incorporate prior knowledge to increase search efficiency while maintaining the black-box character, and thus, the flexibility of our approach?
		\item\textit{{Local Optima:}} How can we avoid converging in a local optimum and discovering the same hazard repeatedly while missing other, perhaps more critical hazards?
	\end{itemize}
	Finally, it should be stressed that our approach is primarily intended to raise awareness of potential hazards. It cannot give a guarantee that a system is safe, because due to the large search space, an exhaustive search would be infeasible. This limitation is part of the fundamental trade-off between the level of detail of the analysis and its exhaustiveness \cite{Vicentini2019}. While a simulation-based approach may miss hazards due to the large search space, a formal approach may miss hazards due to modeling simplifications that are made to keep system complexity manageable. In the long term, it would be desirable to combine simulation-based and formal analysis techniques to combine both methods' strengths.
	
	\section*{ACKNOWLEDGMENT}
	The authors thank Tamim Asfour for fruitful discussions and valuable thought impulses.

	%%%%%%%%%%%%%%%%%%%%%%%%%%%%%%%%%%%%%%%%%%%%%%%%%%%%%%%%%%%%%%%%%%%%%%%%%%%%%%%%
	\bibliography{IEEEabrv,ICRA2021_preprint}
	\bibliographystyle{IEEEtran}
	
\end{document}